\documentclass[letterpaper, 10 pt, conference]{ieeeconf}  
\usepackage{amsmath, amssymb, cite}
\usepackage{makecell}
\usepackage{colortbl}
\usepackage{bbding}
\usepackage{float}
\usepackage{utfsym}
\usepackage{booktabs}
\usepackage{graphicx} 
\usepackage{bm}        
\newcommand{\vect}[1]{\bm{#1}} 
\IEEEoverridecommandlockouts                              
\usepackage{amsmath}
\usepackage{algorithm}
\usepackage[noend]{algpseudocode}
\usepackage{graphicx}
\usepackage{stfloats}
\usepackage{subfigure}
\usepackage{caption}
\usepackage{setspace}
\usepackage[colorlinks,linkcolor=blue]{hyperref}

\usepackage{lipsum}

\newcommand\blfootnote[1]{%
\begingroup 
\renewcommand\thefootnote{}\footnote{#1}%
\addtocounter{footnote}{-1}%
\endgroup 
}

\title{\LARGE \bf
\textbf{STAMICS}: \textbf{S}plat, \textbf{T}rack \textbf{A}nd \textbf{M}ap with \textbf{I}ntegrated \textbf{C}onsistency and {S}emantics for Dense RGB-D SLAM}

\author{Yongxu Wang$^{1*}$, Xu Cao$^{2*}$, Weiyun Yi$^{3*}$,Zhaoxin Fan$^{\dag}$} 

\begin{document}


\twocolumn[{%
\renewcommand\twocolumn[1][]{#1}%
\maketitle

}]

\thispagestyle{empty}
\pagestyle{empty}

\blfootnote{$^{1}$University of Science and Technology of China, Hefei, China.}
\blfootnote{$^{2}$University of Science and Technology Liaoning, Niaoning, China.}
\blfootnote{$^{3}$Beijing Institute of Technology, Beijing, China.}
\blfootnote{$^{\dag}$Beijing Advanced Innovation Center for Future Blockchain and Privacy Computing,
School of Artificial Intelligence, Beihang University, Beijing, China.}


\begin{abstract}
Simultaneous Localization and Mapping (SLAM) is a critical task in robotics, enabling systems to autonomously navigate and understand complex environments. Current SLAM approaches predominantly rely on geometric cues for mapping and localization, but they often fail to ensure semantic consistency, particularly in dynamic or densely populated scenes. To address this limitation, we introduce STAMICS, a novel method that integrates semantic information with 3D Gaussian representations to enhance both localization and mapping accuracy. STAMICS consists of three key components: a 3D Gaussian-based scene representation for high-fidelity reconstruction, a graph-based clustering technique that enforces temporal semantic consistency, and an open-vocabulary system that allows for the classification of unseen objects. Extensive experiments show that STAMICS significantly improves camera pose estimation and map quality, outperforming state-of-the-art methods while reducing reconstruction errors. Code will be public available.
\end{abstract}

\section{INTRODUCTION}
{Simultaneous Localization and Mapping (SLAM)} is a crucial technology in fields such as autonomous driving, robotics, and augmented reality, enabling systems to perceive, map, and navigate complex environments in real time\cite{slamreview1,slamreview2}. The ability to accurately localize and construct detailed maps is fundamental for autonomous systems to interact safely and effectively with their surroundings. As these systems become more integrated into everyday life, the demand for SLAM solutions \cite{segslamreview2,segslamreview3} that are not only geometrically accurate but also semantically rich has grown, especially in high-density, dynamic environments.\cite{fu2024humanplus}

Traditionally, SLAM algorithms \cite{cadena2016past,campos2021orb,engel2014lsd,whelan2015real} have relied heavily on {geometric information} extracted from RGB and depth data to perform localization and mapping. Recent methods, like {Gaussian Splatting} and neural implicit representations, have advanced the field by improving the density and fidelity of scene reconstructions. For instance, {SplaTAM} \cite{splatam} uses explicit volumetric representations for high-quality scene reconstructions, while {SNI-SLAM} \cite{sni} employs implicit neural models for surface reconstruction and semantic labeling. However, a major limitation of these methods is their inability to maintain {semantic consistency} over time. In environments with dense or evolving semantics, this results in {semantic drift}—where the same object may be labeled inconsistently across different time frames—undermining both the accuracy of the mapping process and the utility of the reconstructed scene.

\begin{figure}
    \centering
    \includegraphics[width=1\linewidth]{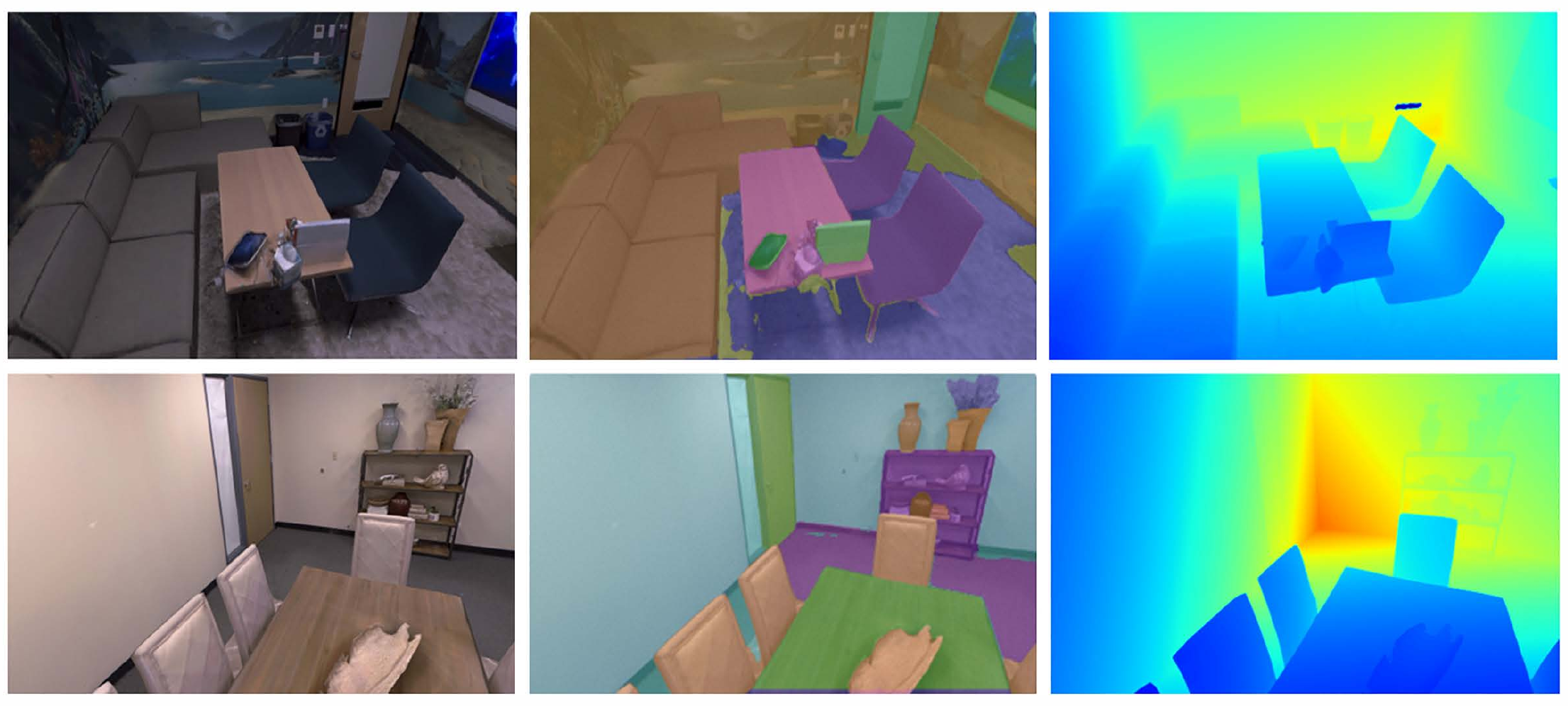}
     \vspace{-0.2in}
    \caption{Illustration of our motivation: From left to right, they are the 3D reconstruction map, the semantic reconstruction map, and the depth reconstruction map.}
    \label{fig:shoutu}
    \vspace{-0.2in}
\end{figure}

To address this issue, we propose {STAMICS}, a novel framework that integrates semantic information directly into the SLAM process through Gaussian Splatting, where semantic data acts as a conditional constraint on geometric reconstruction. As shown in Fig.~\ref{fig:shoutu}, we process semantic information and integrate it into the 3D reconstruction process. Our key idea is to introduce {temporal semantic consistency constraints} that ensure the same object is labeled consistently across time, thereby preventing semantic drift and improving the overall coherence of the system.

The {STAMICS} framework integrates three key components to enhance SLAM performance by maintaining semantic consistency. First, {Semantic-Enhanced Gaussian Splatting} combines semantic information with 3D Gaussian splatting to influence the geometric reconstruction process, ensuring that semantics are preserved and correctly aligned with the geometry of the scene. This improves the system's overall scene understanding and ensures semantic consistency within each frame. Second, {Temporal Semantic Consistency} introduces constraints that ensure consistent semantic labeling of objects across different time frames, preventing semantic drift and maintaining coherence over time. This component enforces the temporal alignment of semantic labels, preventing inconsistencies as the scene evolves. Finally, {Open Vocabulary Expansion} allows the system to dynamically expand its semantic understanding by classifying and labeling previously unseen objects, ensuring consistent semantics even for novel objects and making the system more adaptable to diverse and complex real-world environments. Through extensive experiments, we demonstrate that {STAMICS} achieves state-of-the-art performance in both mapping accuracy and camera pose estimation, significantly reducing reconstruction errors and advancing the robustness of dense SLAM systems.
{Our contributions} can be summarized as follows: 
\begin{itemize}
    \item We propose a novel integration of semantic information into the Gaussian Splatting process to improve geometric reconstruction and maintain semantic consistency within each frame.
    \item We introduce temporal semantic consistency constraints to prevent semantic drift, ensuring robust and coherent mapping over time.
    \item We incorporate an open vocabulary mechanism to handle previously unseen objects, enhancing the system's adaptability and maintaining consistent labeling in real-world scenarios.
\end{itemize}

\section{RELATED WORKS}
Our work is closely related to recent advancements in SLAM technology, particularly in enhancing precision, flexibility, and usability. Therefore, we focus on reviewing two key approaches: SLAM based on Gaussian Splatting and SLAM based on Semantic Injection.

\subsection{SLAM based on Gaussian Splatting} 

With the emergence of Neural Radiance Fields (NeRF), there has been a growing interest in NeRF-based SLAM methods \cite{ibd,lu,rosinol2023nerf,mildenhall2021nerf,zhu2024nicer}. But to address real-time constraints, recent research has shifted towards Gaussian Splatting-based SLAM\cite{kerbl20233d,keselman2022approximate,keselman2023flexible} methods. 3D Gaussian Splatting (3DGS) represents a promising approach for 3D scene modeling, where scenes are represented as a set of 3D Gaussian points, each characterized by parameters such as position, anisotropic covariance, opacity, and color. SplaTAM \cite{splatam} was the first to leverage an explicit volumetric representation of 3D Gaussian distributions, achieving high-fidelity scene reconstruction using an unlocalized RGB-D camera. Other methods, such as Photo-SLAM \cite{photoslam}, combine explicit geometric features with implicit photometric characteristics, learning multi-level features through a Gaussian pyramid-based training method, thereby enhancing the realism of the reconstructed scenes. 

While these methods have significantly advanced the fields of computer vision and SLAM—improving real-time performance, accuracy, and visual realism—they primarily focus on geometric reconstruction and often lack semantic integration. This can lead to semantic drift in complex environments, where objects are labeled inconsistently over time. Our work, STAMICS, addresses these limitations by integrating semantic information directly into the Gaussian Splatting process.

\subsection{SLAM based on Semantic Injection}

Semantic-aware SLAM systems go beyond traditional mapping by not only constructing a 3D map of the environment but also recognizing and understanding the semantic information of objects and regions within a scene. For instance, SNI-SLAM \cite{sni} leverages neural implicit representations to achieve precise semantic mapping, high-quality surface reconstruction, and robust camera tracking. It introduces hierarchical semantic representations, allowing for multi-level semantic understanding of the scene. Similarly, SGS-SLAM \cite{sgs} is a semantic visual SLAM system that utilizes Gaussian rendering to address the over-smoothing issues often present in neural implicit SLAM systems. 

While these methods have advanced the integration of semantics into SLAM\cite{bloesch2018codeslam,salas2013slam++,rosinol2020kimera}, they often face challenges with semantic drift and maintaining temporal semantic consistency in dynamic environments. Our work, STAMICS, tackles these issues by introducing temporal semantic consistency constraints, ensuring stable and coherent semantic labeling over time.

\section{METHOD}
\newcommand{\concat}{\oplus}  
\newcommand{\TD}{\mathrm{TransformerDecoder}}
\newcommand{\MLP}{\mathrm{MLP}}

\begin{figure*}

\vspace{-0.2in}
    \centering
    \includegraphics[width=1\linewidth]{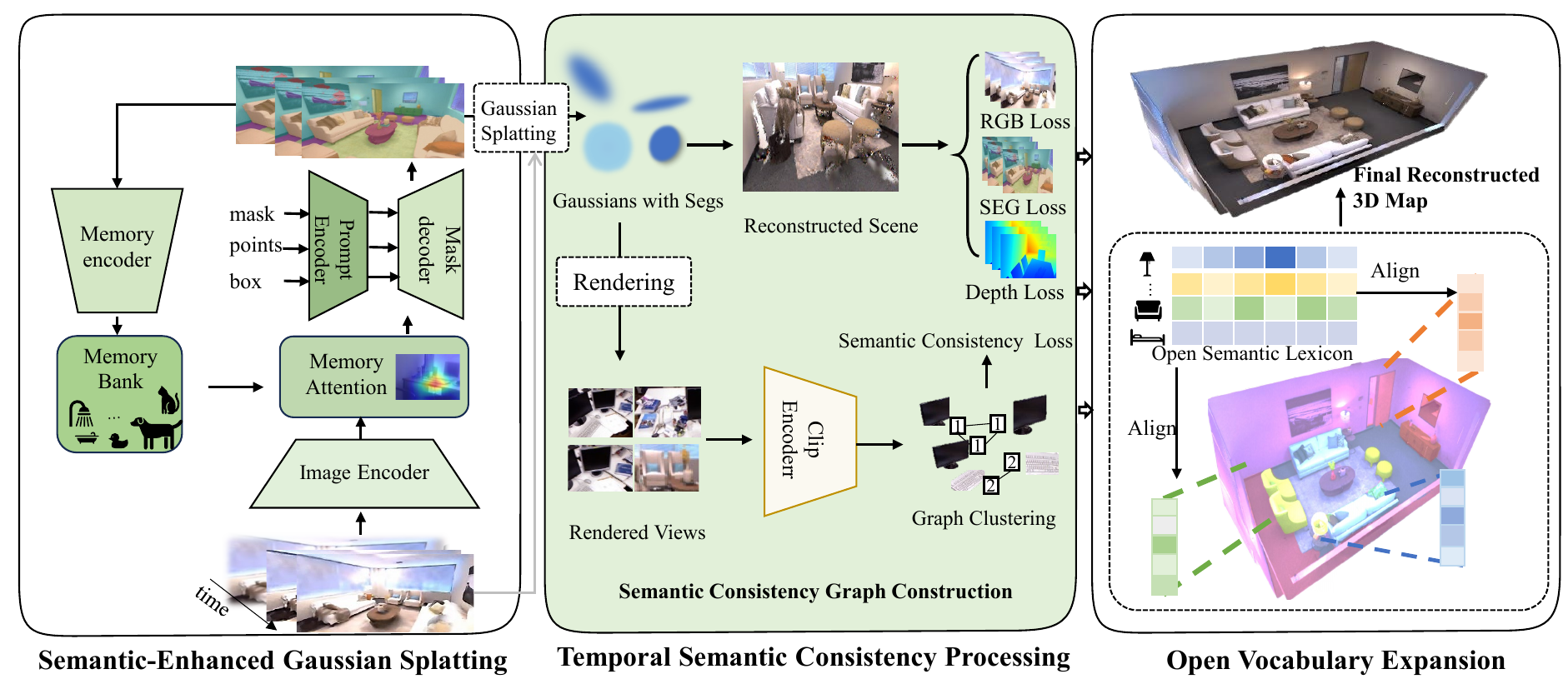}
    \vspace{-0.2in}
    \caption{Overview: The RGB data is processed by the SAM to extract semantic information, which is then fed into the tracking module to localize the camera. Semantic-Enhanced Gaussian Splatting integrates the semantic data into the geometric reconstruction process, ensuring consistency between semantics and geometry. The process is governed by four types of losses, among which the semantic consistency loss originates from the semantic consistency module. The final output features open-vocabulary characteristics, with open vocabulary expansion enabling dynamic learning of new objects and achieving superior reconstruction results.}
    \vspace{-0.2in}

    \label{fig:zhutu}
\end{figure*}

\subsection{Overview}

Given a sequence of images, our objective is to achieve accurate 3D scene reconstruction and robust camera tracking with integrated semantic understanding. To this end, we propose {STAMICS}, a framework that enhances traditional Gaussian-Splatting-based SLAM by incorporating semantic information. The framework is illustrated in Fig.\ref{fig:zhutu}. 

In Gaussian-based SLAM, 3D scene modeling is achieved through parameterized Gaussian distributions. Let $\vect{\mu} \in \mathbb{R}^3$ denote the Gaussian center position, $r$ the radius, $\sigma$ the opacity, and $c$ the color. The 3D Gaussian function is defined as:
\begin{equation}
f^{3D}(\vect{x}) = \sigma \exp \left( -\frac{\|\vect{x} - \vect{\mu}\|^2}{2r^2} \right)
\label{eq:3d_gaussian}
\end{equation}
where $\vect{x} = (x,y,z)^\top$ denotes spatial coordinates. The projection of these Gaussians onto the 2D rendering plane is governed by the camera's pose and intrinsic parameters:
\begin{equation}
\vect{\mu}^{2D} = \vect{K} \frac{\vect{E}_t \vect{\mu}}{d}, \quad 
r^{2D} = \frac{f r}{d}, \quad 
d = (\vect{E}_t \vect{\mu})_z
\label{eq:projection}
\end{equation}

where $f$ denotes the focal length of the camera, $d$ is the depth, 
$\vect{E}_t$ is the extrinsic parameters and $\vect{K}$ represents  intrinsic matrix.
The rendering process is performed by ordering Gaussians by depth and applying front-to-back volumetric rendering. The RGB and depth reconstruction errors for each pixel $p=(u,v)$ are computed as:
\begin{equation}
C(p) = \sum_{i=1}^{n} c_i f_i(p) \prod_{j=1}^{i-1} (1 - f_j(p)) 
\end{equation}
\begin{equation}
D(p) = \sum_{i=1}^{n} d_i f_i(p) \prod_{j=1}^{i-1} (1 - f_j(p)) 
\end{equation}
which is minimized by using differentiable rendering, then optimizing the camera pose while keeping Gaussian parameters fixed.

STAMICS builds upon this foundation and addresses the limitations of traditional Gaussian-Splatting SLAM by incorporating semantic information through three key components: Semantic-Enhanced Gaussian Splatting, which integrates semantic data into the geometric reconstruction process to ensure alignment between semantics and geometry; Temporal Semantic Consistency, which introduces temporal constraints to maintain consistent semantic labeling across frames, mitigating semantic drift; and Open Vocabulary Expansion, which enables dynamic learning of new objects, expanding the system's semantic understanding to handle diverse environments. In the following sections, we will delve into each of these components.
\subsection{Semantic-Enhanced Gaussian Splatting}
Semantic information is often overlooked in traditional Gaussian-based SLAM systems, limiting the system’s ability to interpret scenes with nuanced understanding. To address this, we first extract semantic features from the input images and inject them into the Gaussians.

Scene Semantic Extraction: The process begins with an image encoder that translates visual inputs into high-dimensional feature representations. These features are then processed by a memory attention module, which leverages a memory bank containing historical data. This module enriches the feature representations by emphasizing relevant details, ensuring that critical information from past observations is retained. Concurrently, a prompt encoder generates task-specific cues, guiding a mask decoder to segment and extract precise semantic regions from the image. Once the image has been segmented, the semantic information is re-encoded by a memory encoder and fed back into the memory bank. This continuous feedback loop allows the system to dynamically evolve its understanding of the environment, progressively refining its semantic knowledge over time. The entire process can be encapsulated as:

\begin{equation}
\begin{aligned}
M_t &= f_{\text{mem}} \Big( W_D \cdot f_{\text{mask}} \Big( W_P \cdot f_{\text{prompt}} \Big( \\
& \quad A \Big( W_A \cdot f_{\text{att}}(E(I_t), M_{t-1}) \Big) \Big) \Big) \Big)
\end{aligned}
\end{equation}

where \( I_t \in \mathbb{R}^{H \times W \times 3} \) is the input image at time \( t \), \( F_t = E(I_t) \in \mathbb{R}^{d_f} \) is the feature embedding from the image encoder, \( A(F_t, M_{t-1}) \in \mathbb{R}^{d_f} \) is the attention mechanism using memory \( M_{t-1} \), \( W_A \in \mathbb{R}^{d_f \times d_a} \), \( W_P \in \mathbb{R}^{d_a \times d_p} \), and \( W_D \in \mathbb{R}^{d_p \times d_m} \) are learned weight matrices, \( f_{\text{att}}, f_{\text{prompt}}, f_{\text{mask}}, f_{\text{mem}} \) are non-linear activation functions, and \( M_t \in \mathbb{R}^{d_m} \) is the updated memory representation at time \( t \).

{Semantic Injection}: After the semantic features are extracted, they are integrated into the Gaussian splatting process through \textit{Semantic Injection}. This allows the system to incorporate both geometric and semantic information into the scene reconstruction and tracking processes. Each Gaussian, traditionally parameterized by position, radius, opacity, and color, is now augmented to carry semantic information. Specifically, each Gaussian is represented as a vector $c_i = [r_i, g_i, b_i, seg_i]^T$, where $r_i$, $g_i$, and $b_i$ represent the color channels, and $seg_i$ encodes the semantic label associated with the Gaussian. This enriched representation ensures that both visual and semantic contexts are accounted for during scene reconstruction. To ensure semantic consistency during rendering, we define the semantic reconstruction error for each pixel $p = (u, v)$ as:

\begin{equation}
S(p) = \sum_{i=1}^{n} s_i f_i(p) \prod_{j=1}^{i-1} (1 - f_j(p))
\end{equation}

Here, $s_i$ represents the semantic value of the $i$-th Gaussian, and $f_i(p)$ is the contribution of the $i$-th Gaussian to the pixel $p$, based on its position and depth. This formulation uses front-to-back volumetric rendering to ensure that the semantics are consistently projected onto the 2D image plane.

\subsection{Temporal Semantic Consistency Processing}

Incorporating semantic information from 2D masks improves instance recognition but lacks consistency across frames\cite{openmask3d,maskclustering}. To address this, we propose a Temporal Semantic Consistency Processing pipeline to enforce instance consistency over time. Our approach aligns instances across frames using three key components: Semantic Consistency Graph Construction, Graph Clustering, and Semantic Consistency Loss. First, we construct a semantic consistency graph with latent features from 2D masks to find correspondences across frames. Next, graph clustering groups nodes (masks) representing the same instance. Finally, we compute a semantic consistency loss to refine the 3D Gaussian mapping by aligning their projected semantic properties with 2D semantic maps. Next, we detail the process.

\begin{figure}[ht]
    \centering
    \includegraphics[width=1\linewidth]{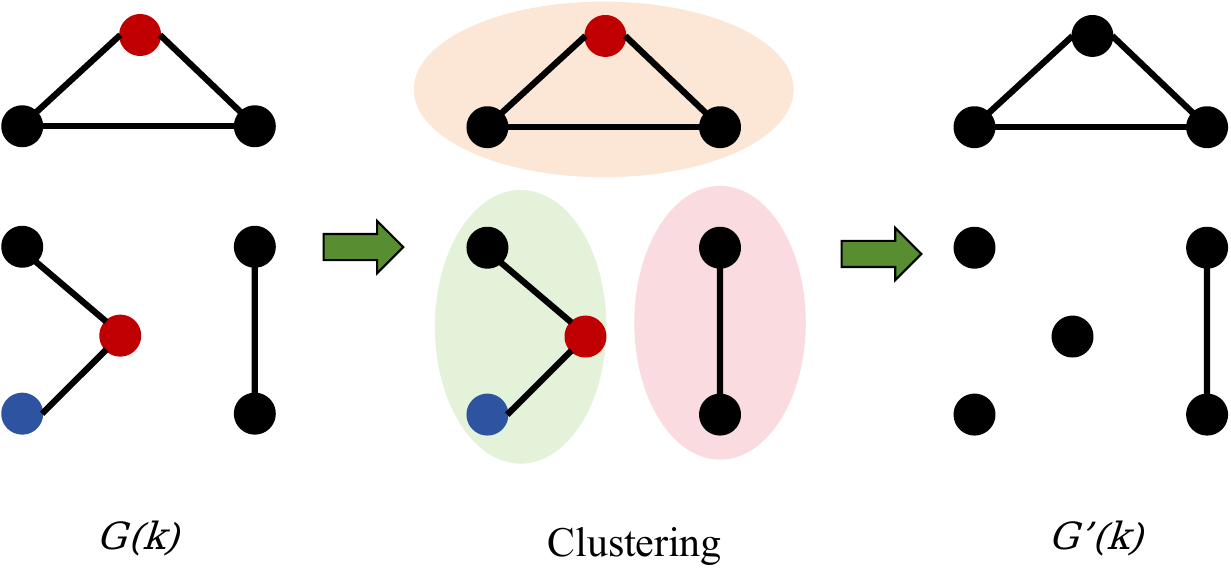}
    \vspace{-0.2in}
   
    \caption{Illustration of graph clustering. Nodes with high semantic consistency scores are grouped into the same category in the graph \(G(k)\). Edges between inconsistent nodes are removed, resulting in a new graph \(G'(k)\).}
    \vspace{-0.1in}
    \label{fig:graph-clustering}
\end{figure}

{Semantic Consistency Graph Construction:} We first construct the semantic consistency graph to cluster 2D masks from different frames that likely represent the same instance. Each 2D mask is treated as a node, with latent features \(F_i\) extracted and assigned to node \(i\). For each pair of nodes \(i\) and \(j\) from different frames, we compute the cosine similarity between their latent features, and an edge is formed if the similarity exceeds a predefined threshold \(\tau\). The edge function \(E(i, j)\) is defined as:

\begin{equation}
E(i, j) = \mathbb{I}\left( \frac{F_i \cdot F_j}{\|F_i\| \|F_j\|} \geq \tau \right)
\end{equation}
where \(\mathbb{I}(\cdot)\) is the indicator function that returns 1 if the condition is true and 0 otherwise. In our case, we set \(\tau = 0.8\). This process is repeated for all nodes across frames within a sliding window, as shown in Fig.~\ref{fig:graph-clustering}, forming the edges of the semantic consistency graph, which captures potential correspondences of instances across frames.

\begin{figure}[ht]
    \centering
    \includegraphics[width=1\linewidth]{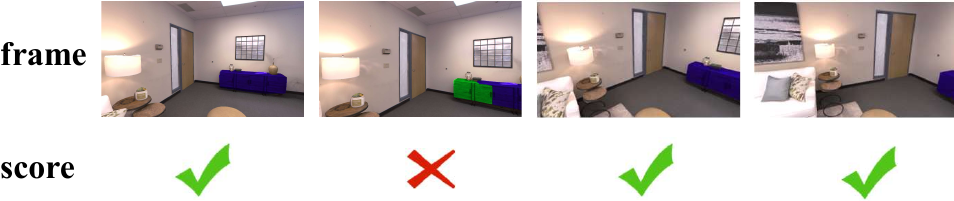}
    \vspace{-0.2in}
    \caption{Illustration of the consistency score. The semantic label for the cabinet in the second frame is inconsistent. For the cabinet node in the first frame, the consistency score is \(3/4\).}
    \vspace{-0.1in}
    \label{fig:seg-score}
\end{figure}

{Graph Clustering:} Once the graph is constructed, nodes are clustered to determine whether 2D masks from different frames correspond to the same instance. For each node \(i\), the semantic consistency score \(S_i\) is calculated as:

\begin{equation}
S_i = \frac{\sum_{j \in N_i} \delta(L_i, L_j)}{|N_i|}
\end{equation}

where \(\delta(L_i, L_j)\) equals 1 if labels match, and 0 otherwise. Nodes with a score greater than \(\frac{2}{3}\) are grouped into the same cluster, thereby updating the semantic labels across frames, as illustrated in Fig.~\ref{fig:seg-score}.

{Semantic Consistency Loss:} Finally, we compute the semantic consistency loss using the updated 2D semantic frames and the initialized 3D Gaussian parameters. Let \(S_{\text{splat}}^{(f)}(x, y)\) represent the splatted 2D semantic frame from the initialized Gaussians for frame \(f\) at pixel \((x,y)\), and let \(S_{\text{updated}}^{(f)}(x, y)\) be the updated frame. The semantic consistency loss \(L_{\text{sc}}\) over \(N\) frames is given by:

\begin{equation} 
L_{\text{sc}} = \sum_{f=1}^{N} \sum_{(x, y)} \left| S_{\text{splat}}^{(f)}(x, y) - S_{\text{updated}}^{(f)}(x, y) \right|
\end{equation}

This loss ensures that the 3D Gaussians maintain semantic alignment with the updated 2D projections across frames, reinforcing temporal instance consistency.

\subsection{Open-vocabulary Expansion}
\label{sec:open_feature}
To enhance the representation capability in the Temporal Semantic Consistency Processing and dynamically expand the system's semantic understanding by classifying and labeling previously unseen objects, we introduce an Open-vocabulary Expansion. This ensures consistent semantics even for novel objects, making the system more adaptable to diverse and complex real-world environments. Specifically, we leverage open-vocabulary features to move beyond predefined label sets, enabling the identification and differentiation of a wider variety of instances. This allows the system to flexibly represent diverse objects and maintain semantic consistency across frames. 

Our approach employs the Segment Anything Model (SAM) \cite{SAM} to generate mask regions on the original RGB images, followed by the Contrastive Language–Image Pre-training (CLIP) model \cite{CLIP} to extract open-vocabulary features. Formally, given an image \( I \) and its corresponding mask \( M \), the open-vocabulary feature \( f \) is computed as:
\begin{equation}
f = \mathcal{F}_{\text{CLIP}}\left( \mathcal{T}( \mathcal{F}_{\text{SAM}}(I) \odot M ) \right)
\end{equation}
where \( \mathcal{F}_{\text{SAM}}(I) \) denotes the feature map generated by the SAM model, \(\odot\) represents the element-wise multiplication with the mask \(M\) to isolate the region of interest, and \( \mathcal{T}(\cdot) \) is a transformation operation (e.g., cropping or resizing) applied to the masked region. Finally, \( \mathcal{F}_{\text{CLIP}}(\cdot) \) extracts the high-dimensional open-vocabulary feature from the transformed image patch.

In practice, we use this open-vocabulary feature as the latent feature in the Temporal Semantic Consistency Processing. These latent features serve as the basis for constructing the semantic consistency graph, where nodes represent the extracted features for each instance across frames. The graph is then clustered to group nodes that likely correspond to the same instance. By using these open-vocabulary latent features, we ensure that our method can dynamically handle diverse and previously unseen objects, maintaining instance consistency and improving the robustness of the graph construction and clustering processes.

\subsection{Semantic Consistency-Based Gaussian Map Refinement}
After achieving semantic consistency through Temporal Semantic Consistency Processing, we focus on optimizing the Gaussian parameters and improving tracking performance. This is achieved via a combination of differential rendering and gradient-based optimization, where the camera poses are fixed, and the Gaussians are updated to refine the scene representation.

The core optimization problem is formulated as minimizing a combined loss function that integrates RGB, depth, and semantic information.
\begin{equation}
\begin{aligned}
L_t = \sum_{p} (\mathbf{1}(s(p) > 0.99)) (\\
L_1(D(p)) + 0.5L_1(C(p))+ 1.5L_1(S(p)))
\end{aligned}
\end{equation}

where \( C(p) \), \( D(p) \), and \( S(p) \) represent the RGB, depth, and semantic projections at pixel \( p \), respectively. The weighting factors are set as \( w_{\text{rgb}} = 0.5 \), \( w_{\text{depth}} = 1 \), and \( w_{\text{semantic}} = 1.5 \) to balance visual details, spatial structure, and semantic understanding.

Given the semantic consistency established across frames, we further optimize the Gaussians by selecting keyframes that maximize overlap. For each current frame \( f_{\text{cur}} \), we project its depth map into a 3D point cloud \( P_{\text{cur}} \) and compute the overlap with previous keyframes \( f_k \):

\begin{equation}
\text{Overlap}(f_{\text{cur}}, f_k) = \frac{|P_{\text{cur}} \cap P_k|}{|P_{\text{cur}}|}
\end{equation}

We select the top \( k \) keyframes with the highest overlap to update the Gaussian parameters.

To ensure temporal consistency in the Gaussian map, we employed the semantic consistency
loss $L_{\text{sc}}$. The overall optimization problem thus becomes:
\begin{equation}
L_{\text{opt}} = L_{\text{total}} + 2L_{\text{sc}}
\end{equation}

By minimizing \( L_{\text{opt}} \), we achieve precise updates to the Gaussian parameters, ensuring that both geometric and semantic information is accurately captured and optimized in highly dynamic environments.

\section{EXPERMENT}

\begin{figure*}

\vspace{-0.2in}
    \centering
    \includegraphics[width=1\linewidth]{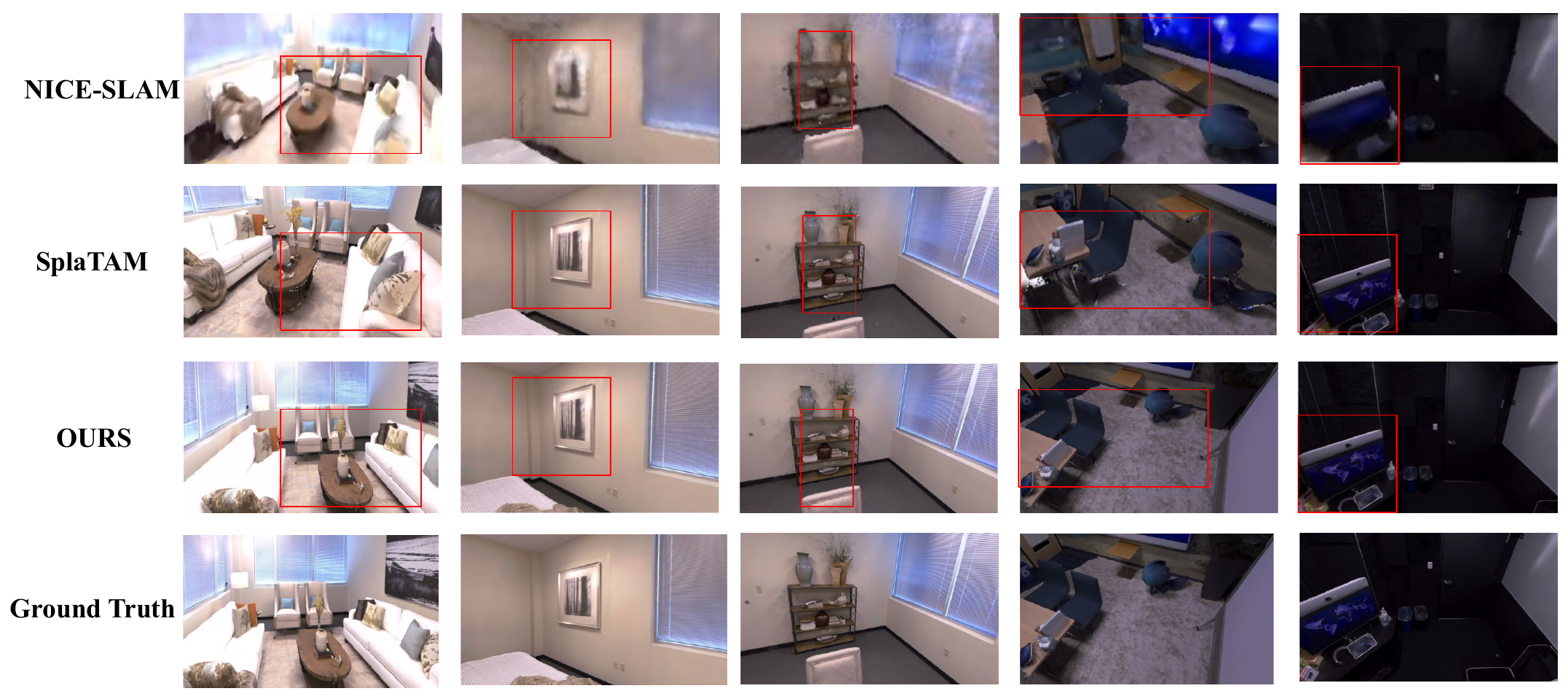}
    \vspace{-0.2in}
    \caption{Comparison of reconstruction results with existing methods}
    \vspace{-0.2in}
    \label{fig:enter-label}
\end{figure*}

\subsection{Implementation Details}

We conduct experiments on three benchmark datasets: TUM-RGBD \cite{TUM}, Replica \cite{straub2019replica}, and Scannet \cite{scannet}. The Replica dataset consists of high-quality synthetic indoor environments, providing accurate depth maps and minimal inter-frame camera pose changes, making it suitable for evaluating performance under ideal conditions. In contrast, the TUM-RGBD dataset poses significant challenges for dense SLAM methods due to its low-quality RGB and depth images, sparse and noisy depth maps, and frequent motion blur. The Scannet dataset, similar to TUM-RGBD \cite{TUM}, is used for experiments with evaluations performed every five frames. To assess the influence of semantic consistency on mapping performance, we perform ablation studies on TUM-RGBD \cite{TUM}. Performance is evaluated using standard metrics. For RGB image rendering, we employ Peak Signal-to-Noise Ratio (PSNR), Structural Similarity Index (SSIM), and Learned Perceptual Image Patch Similarity (LPIPS), each examining different aspects of image quality. Depth map accuracy is measured using L1 loss. Camera pose estimation accuracy is assessed using the Root Mean Square Error of the Absolute Trajectory Error (ATE RMSE). Additionally, we introduce a segmentation loss (seg loss) to quantify semantic consistency errors. We compare our method against several state-of-the-art approaches. The primary baseline is SplaTAM \cite{splatam}, a leading 3D Gaussian SLAM technique. We also include comparisons with advanced SLAM methods based on dense radiance fields, including Point-SLAM \cite{point}, NICE-SLAM \cite{nice}, and Vox-Fusion \cite{vox}, to provide a comprehensive evaluation of our approach.

\begin{table}[ht!]

\centering
\resizebox{\linewidth}{!}{
\begin{tabular}{ccccccc}
\toprule
\textbf{Methods}   & \textbf{Avg.} & \textbf{desk}& \textbf{desk2}& \textbf{room}& \textbf{xyz}& \textbf{off}\\ \midrule
Kintinuous \cite{whelan2012kintinuous} & \textbf{4.84}        & 3.70              & 7.08               & \textbf{7.50}& 2.93             & 2.98              \\
ElasticFusion \cite{whelan2015elasticfusion} & 6.90          & \textbf{2.52}& 6.81               & 21.48             &\textbf{1.16}&\textbf{2.53}              \\
NICE-SLAM \cite{nice} & 15.86         & 4.24              &\underline{4.98}               & 34.48             & 31.76            & \underline{3.85}              \\
Vox-Fusion \cite{vox} & 11.32         & 3.53              & 6.01              & 19.54            & 1.48             & 26.04\\
Point-SLAM \cite{point} & 8.70          & 4.34              & \textbf{4.54}               & 29.82             & 1.33             & 3.48              \\
SplaTAM \cite{splatam}            & 5.53          & 3.38              & 6.60               &11.13             & 1.38             & 5.16              \\
 ours& \underline{5.11}&\underline{3.22}& 5.57& \underline{10.62}&\underline{1.18}&4.98\\ \bottomrule
\end{tabular}}
\vspace{0.0in}
\caption{ATE comparison using TUM-RGBD \cite{TUM} dataset.Best results are highlighted in bold as \textbf{first}, and \underline{second} for the second-best.}
\vspace{-0.15in}
\label{tab:t1}
\end{table}
\subsection{Quantitative Comparison}
We begin by conducting a quantitative analysis of our SLAM method's performance, as shown in Table \ref{tab:t1}, Table \ref{tab:t2}, Table \ref{tab:t3}, Table \ref{tab:t4} and Table \ref{tab:t5}.

\begin{table}[ht!]
\centering
\vspace{-0.03in}
\resizebox{\linewidth}{!}{
\begin{tabular}{@{}ccccccc}
\toprule
\textbf{Methods}   & \textbf{Avg.} & \textbf{r0}& \textbf{r1}& \textbf{of0}& \textbf{of1}& \textbf{of2}\\ \midrule

NICE-SLAM \cite{nice}&1.10 &1.02         & 1.44             & 0.92              & 1.06                       & 1.12              \\
Vox-Fusion \cite{vox}&  3.99        & 1.38             & 4.84               & 8.48             & 2.66            & 2.58\\
Point-SLAM \cite{point}& 0.52 &0.66          & 0.43             &\textbf{ 0.41 }              & 0.52             & 0.56                        \\
SplaTAM \cite{splatam}& \underline{0.39}        &\underline{ 0.38  }        & \textbf{0.38}             & 0.52              &\underline{ 0.34  }           & \underline{0.33}                      \\
 ours& \textbf{0.33} & \textbf{0.25}& \textbf{0.38} & \underline{ 0.44}& \textbf{0.22} & \textbf{0.26}\\ \bottomrule
\end{tabular}}
\caption{ATE comparison using Replica \cite{straub2019replica}dataset.Best results are highlighted in bold as \textbf{first}, and \underline{second} for the second-best.}
\vspace{-0.15in}
\label{tab:t2}
\end{table}

On the TUM-RGBD \cite{TUM} dataset, our approach exhibited exceptional stability, achieving an Avg. ATE of 5.11, outperforming other methods in overall consistency. Specifically, in the 'xyz' sequence, our method recorded an error of 1.18, nearly matching the best-performing ElasticFusion at 1.17, indicating strong adaptability across diverse environments. In the more challenging 'room' sequence, our method significantly outperformed ElasticFusion, with errors of 10.62 and 21.49, respectively, representing a 6\% improvement over previous baselines. These results are detailed in Table \ref{tab:t1}. On the Replica \cite{straub2019replica} dataset, our method achieved the lowest average error of 0.33, outperforming all other methods. In the 'r0' sequence, we achieved an error of 0.25, marking a 15\% improvement over the closest baseline. This highlights our method's robustness in environments characterized by high consistency and repetitive structures, as shown in Table \ref{tab:t2}. On the Scannet \cite{scannet} dataset, our method matched the performance of NICE-SLAM \cite{nice}, achieving an average error of 9.68. In the '0000' sequence, our method further distinguished itself with an error of 5.36, significantly outperforming NICE-SLAM (12.68) and Vox-Fusion (59.98), reflecting a 15\% improvement over baseline methods (Table \ref{tab:t3}). This demonstrates our method's reliability in complex indoor settings. 

\begin{table}[ht!]
\vspace{-0.05in}
\centering
\resizebox{\linewidth}{!}{
\begin{tabular}{@{}ccccccc}
\toprule
\textbf{Methods}   & \textbf{Avg.} & \textbf{0000}& \textbf{0106}& \textbf{0169}& \textbf{0181}& \textbf{0201}\\ \midrule
NICE-SLAM \cite{nice} & \underline{9.69}         &\underline{12.68}              & \underline{8.02}               &\textbf{10.88}             & 12.98            &\textbf{3.87}              \\
Vox-Fusion \cite{vox} &26.00          & 59.98              & 9.18               & 26.69             & 24.46             & 9.68\\
Point-SLAM \cite{point} & 10.70         & 10.26             &\textbf{ 7.92 }              &\textbf{10.88}             & 14.89            & 9.56            \\
SplaTAM \cite{splatam}         &   12.32        & 12.86             & 17.86               & 12.10             &\underline{ 11.22 }            & 7.56              \\
 ours&\textbf{9.68} &\textbf{5.36} &15.40 &11.49 &\textbf{8.80} &\underline{7.34}\\  \bottomrule
\end{tabular}}
\vspace{-0.05in}
\caption{ATE comparison using Scannet \cite{scannet} dataset.Best results are highlighted in bold as \textbf{first}, and \underline{second} for the second-best.}
\vspace{-0.15in}
\label{tab:t3}
\end{table}

Beyond trajectory accuracy, our method also excels in rendering quality. In the 'R1' scenario, we achieved the PSNR of 38.53 and the SSIM of 0.98, substantially outperforming Vox-Fusion’s PSNR of 27.79 and SSIM of 0.86 (Table \ref{tab:t4}). In dynamic environments, such as 'Of0' and 'Of1', our method maintained high PSNR values (38.26 and 39.28) and achieved low LPIPS scores (0.10 and 0.09), demonstrating robustness in challenging, dynamic conditions. Compared to SplaTAM \cite{splatam}, our method consistently delivered superior image quality and depth estimation. As shown in Table \ref{tab:t5}, our method achieved an average PSNR of 23.36, exceeding SplaTAM’s 22.21, with notable improvements in the 'xyz' sequence (26.68 vs. 25.15). These results validate the robustness and generalization capabilities of our method, especially in complex environments.

\begin{table}[ht]
\centering
\resizebox{\linewidth}{!}{
\begin{tabular}{@{}ccccccc}
\hline 
\textbf{Methods} & \textbf{Metrics} & \textbf{Avg.} & \textbf{R0} & \textbf{R1} & \textbf{Of0} & \textbf{Of1} \\ \hline
& PSNR $\uparrow$  & 25.28 & 22.41 & 22.33 & 27.80 & 29.81 \\ 
 \textbf{Vox-Fusion \cite{vox} }& SSIM  $\uparrow$  & 0.85 & 0.69 & 0.76 & 0.84 & 0.91 \\ 
 & LPIPS & 0.26 & 0.32 & 0.28 & 0.26 & 0.17 \\ \hline
& PSNR  $\uparrow$  & 26.00 & 22.11 & 22.46 & 29.08 & 30.35 \\ 
 \textbf{NICE-SLAM \cite{nice}}& SSIM  $\uparrow$  & 0.76 & 0.68 & 0.77 & 0.88 & 0.90 \\ 
 & LPIPS  & 0.34 & 0.34 & 0.28 & 0.21 & 0.19 \\ \hline
 
& PSNR $\uparrow$   & 35.93 & 32.38 & 34.07 & 38.25 & 39.13 \\ 
 \textbf{Point-SLAM \cite{point}}& SSIM $\uparrow$   & 0.97 & \textbf{0.98 }& 0.97 & \textbf{0.99}& \textbf{0.99 }\\ 
 & LPIPS  & 0.12 & 0.11 & 0.12 & 0.10 & 0.16 \\ \hline
& PSNR  $\uparrow$ &36.03 &32.84 & 33.88 & 38.24 & 39.16 
\\ 
 \textbf{SplaTAM \cite{splatam}}& SSIM  $\uparrow$  & \textbf{0.97} &\textbf{ 0.98} & 0.97 & \textbf{0.99} & 0.97 
\\ 
 & LPIPS$\downarrow$ & 0.11 &\textbf{ 0.07} & 0.12 & 0.10 & \textbf{0.09} \\ \hline

 & PSNR $\uparrow$ & \textbf{38.24} & \textbf{36.89} & \textbf{38.53} & \textbf{38.26 }& \textbf{39.28} 
\\ 
 \textbf{Ours}& SSIM $\uparrow$ & \textbf{0.97} & 0.96 & \textbf{0.98} & \textbf{0.99} & 0.98
\\ 
 & LPIPS $\downarrow$ &\textbf{ 0.09} & 0.08 & \textbf{0.10} &\textbf{ 0.09} & \textbf{0.09} \\ \hline
\end{tabular}}
\vspace{-0.05in}
\caption{Comparison of different methods on image quality metrics. he best metrics for PSNR, LPIP, and SSIM are highlighted in \textbf{bold}.}
\vspace{-0.21in}
\label{tab:t4}
\end{table}

In conclusion, our SLAM method consistently outperforms state-of-the-art approaches across multiple datasets and scenarios, delivering superior accuracy in both trajectory estimation and scene reconstruction. It demonstrates strong adaptability, stability, and precision, making it highly effective in a wide range of challenging environments.

\begin{table}[ht!]
\centering
\vspace{-0.1in}
\resizebox{\linewidth}{!}{
\begin{tabular}{@{}ccccccc}
\hline 
\textbf{Methods} & \textbf{Metrics} & \textbf{Avg.} & \textbf{D1} & \textbf{D2} & \textbf{R1} & \textbf{xyz} \\ \hline
& PSNR$\uparrow$ &22.21  & 21.49 & 20.98 & 21.22 & 25.15 
\\ 
& SSIM$\uparrow$  &0.83 & 0.83& 0.79 & 0.82 &0.89 
\\ 
 \textbf{SplaTAM \cite{splatam}} & LPIPS $\downarrow$&0.23  & 0.27 &0.27 & 0.27 & 0.12 \\ 
& Depth L1 $\downarrow$&3.49 &4.96 &3.42 & 3.30 & 2.31 \\ 
& Seg L1 &\XSolidBrush & \XSolidBrush & \XSolidBrush &  \XSolidBrush & \XSolidBrush \\ \hline

 & PSNR $\uparrow$ & \textbf{23.36} & \textbf{23.49} &\textbf{21.04} &\textbf{ 22.23} & \textbf{26.68}
\\ 
& SSIM $\uparrow$ &\textbf{0.87}  &\textbf{ 0.90 }&\textbf{0.81} & \textbf{0.86} &\textbf{ 0.90}
\\ 
 \textbf{Ours} & LPIPS $\downarrow$ &\textbf{ 0.19} & \textbf{0.19} &\textbf{ 0.26} & \textbf{0.23 }& \textbf{0.09} \\ 
 & Depth L1 $\downarrow$ &\textbf{3.04}  & \textbf{3.16} &\textbf{3.42 } &\textbf{ 3.30} & \textbf{2.28 }\\
 & Seg L1 &\textbf{0.49}  & \textbf{0.57} &\textbf{ 0.50} &\textbf{0.44} & \textbf{0.46}\\ \hline
\end{tabular}}
\vspace{-0.02in}
\caption{Comparison with baseline methods using TUM-RGBD,The improved indicators are marked in \textbf{bold}.}
\vspace{-0.2in}
\label{tab:t5}
\end{table}

\subsection{Qualitative Comparison}
We provide qualitative visual comparisons to further highlight the superiority of our method over existing approaches. As shown in Fig.~\ref{fig:enter-label}, these visualizations demonstrate the enhanced detail, accuracy, and consistency of our SLAM system compared to SplaTAM \cite{splatam} and NICE-SLAM \cite{nice}. The improvements in visual fidelity—particularly in texture and structural reconstruction—reinforce the quantitative gains observed in metrics such as PSNR, SSIM, and LPIPS. Fig. \ref{fig:enter-label} clearly illustrates that our method (OURS) exhibits a distinct advantage in visual reconstruction over other methods (NICE-SLAM \cite{nice} and SPLaTAM \cite{splatam}), with three key areas of improvement:  1) Superior detail preservation: Our approach is more precise in handling intricate details, such as items on a table or frames on a wall, resulting in sharper, more defined edges with minimal blurring or distortion.  2) Higher structural integrity: Our method maintains the geometric shapes and spatial structure of objects with greater accuracy. For instance, the furniture in the scene, such as sofas and chairs, closely resembles the real-world geometry (Ground Truth), whereas other methods exhibit noticeable distortion or deformation. 3) Stronger texture consistency: The textures in our reconstructions are more faithful to the Ground Truth, offering higher resemblance to the real scene.

\subsection{Ablation Study}

Our method is built upon three key components: semantic-enhanced Gaussian splatting, temporal semantic consistency processing, and open-vocabulary expansion. To evaluate the impact of each component, we conduct comprehensive ablation studies to analyze their contributions to the overall performance of the SLAM system.

{Impact of Semantic-Enhanced Gaussian Splatting:} We first investigate the effect of incorporating semantic information into the SLAM system through the loss function. The addition of a semantic module allows the system to process geometric data and capture the semantic structure of the environment. This significantly improves the accuracy of relocalization and loop closure. As shown in Table.\ref{tab:t6}, our experiments show that the ATE decreases substantially compared to the baseline model, which lacks semantic information. Additionally, the semantic module enhances the model's ability to identify and associate environmental features, reducing mismatches and trajectory estimation errors, ultimately optimizing the overall loss function. These results confirm that incorporating semantics into the Gaussian splatting improves both localization accuracy and map quality.

\begin{table}[ht!]
\centering
\vspace{-0.07in}
\resizebox{\linewidth}{!}{
\begin{tabular}{@{}ccccccc@{}}
\toprule
\textbf{Methods}   & \textbf{Avg} & \textbf{desk}& \textbf{desk2}  & \textbf{room}& \textbf{off } \\ \midrule
Without  Seg&5.81 & 3.34        & 6.58            & 11.49   & 5.18                 \\
With Seg&5.27  & 3.28         & 5.58  & 10.82      & 4.69            
\\ \bottomrule

\end{tabular}}
\vspace{-0.05in}
\caption{An ablation study was conducted on the TUM-RGBD dataset to compare the performance before and after incorporating semantic information.}
\vspace{-0.1in}
\label{tab:t6}
\end{table}

{Impact of Temporal Semantic Consistency Processing:} Next, we evaluate the effect of introducing a semantic consistency module to the SLAM system. This module enforces the consistency of semantic information during map construction and feature association, particularly in dynamic or complex scenes. Our experiments indicate that this component significantly improves the alignment between point cloud data and the map, and reduces errors in depth estimation. By maintaining high semantic consistency, the system achieves more accurate context-aware depth estimation, directly reducing depth loss and enhancing overall stability. As shown in Table. \ref{tab:t7}, metrics such as SSIM  and LPIPS show marked improvements, with LPIPS improving by 50\% and depth errors decreasing by 24\% compared to the baseline model. These findings demonstrate the importance of preserving semantic consistency in challenging environments.

{Impact of Open-Vocabulary Expansion:} Finally, we assess the contribution of the open-vocabulary module, which extends the system's recognition capabilities to include previously unseen objects and scenes. This expansion greatly enhances the system's adaptability and flexibility, allowing it to operate effectively in more variable and unknown environments. Experimental results show that the open-vocabulary module improves the system's understanding of complex environments and increases the quality and realism of the reconstructed 3D scenes. Broader semantic recognition also enhances the model's performance in tasks such as depth estimation and environmental understanding. This component significantly boosts the system's overall performance and adaptability, particularly in diverse and dynamic settings.

In summary, the ablation studies clearly demonstrate the contributions of each key component to the system's performance. The introduction of open vocabulary aimed to expand the system's recognition range, enabling it to identify and understand previously unseen objects and scenes. This capability greatly enhanced the model's adaptability and flexibility, allowing it to function effectively in more variable and unknown environments. Experimental results confirmed that the use of open vocabulary not only enhanced the model's understanding of complex environments but also, through broader semantic recognition, further improved the quality and realism of the reconstructed 3D scenes. The incorporation of open vocabulary allowed the model to better perform depth estimation and environmental understanding tasks when facing diverse environments, significantly enhancing the model's overall performance and adaptability.

\begin{table}[ht!]
\centering
\vspace{-0.15in}
\resizebox{\linewidth}{!}{
\begin{tabular}{@{}ccccccc@{}}
\toprule
\textbf{Methods}   & \textbf{PSNR}& \textbf{Depth L1}  & \textbf{SSIM}& \textbf{LPIP} \\ \midrule
Without  Seg        & 21.49            & 4.38   & 0.83& 0.26                     \\
With Seg        & 21.60& 3.34      & 0.84            &0.24                 \\
With Seg Consistency       & 23.54            &\textbf{3.30}
& 0.88          &\textbf{0.13}
 
\\ \bottomrule

\end{tabular}}
\caption{Performance comparison using TUM-RGBD dataset.The indicators with significant improvements are marked in \textbf{bold}.}
\vspace{-0.2in}
\label{tab:t7}
\end{table}

\section{CONCLUSION}
We present STAMICS, a novel SLAM method that integrates semantic information with 3D Gaussian representations to improve localization and mapping accuracy. By combining a 3D Gaussian-based scene representation for high-fidelity reconstruction, a graph-based clustering technique to enforce temporal semantic consistency, and an open-vocabulary system for classifying unseen objects, STAMICS achieves significant improvements in camera pose estimation and map quality. Extensive experiments demonstrate that our approach outperforms state-of-the-art methods while reducing reconstruction errors. 

\addtolength{\textheight}{-12cm}   







\bibliographystyle{IEEEtranS}
\bibliography{main}
\end{document}